\documentclass[10pt,twocolumn,letterpaper]{article}

\usepackage{cvpr}              

\usepackage{graphicx}
\usepackage{amsmath}
\usepackage{amssymb}
\usepackage{booktabs}
\usepackage{algorithm}
\usepackage{algpseudocode}
\usepackage{float}
\usepackage[english]{babel}

\usepackage[pagebackref,breaklinks,colorlinks]{hyperref}

\usepackage[capitalize]{cleveref}
\crefname{section}{Sec.}{Secs.}
\Crefname{section}{Section}{Sections}
\Crefname{table}{Table}{Tables}
\crefname{table}{Tab.}{Tabs.}

\def\cvprPaperID{*****} 
\def\confName{CVPR}
\def\confYear{2023}

\begin{document}

\title{DeepSegmenter: Temporal Action Localization for Detecting Anomalies in Untrimmed Naturalistic Driving Videos}

\author{ \thanks{Corresponding and primary author} Armstrong Aboah \& Ulas Bagci\\
Department of Radiology\\
Northwestern University\\
{\tt\small {armstrong.aboah, ulas.bagci}@northwestern.edu}
\and
 Abdul Rashid Mussah \thanks{equal contribution} \& Neema Jakisa Owor \thanks{equal contribution}\\
Department of Civil Engineering\\
University of Missouri-Columbia\\
{\tt\small {akm2fx,nodyv,}@umsystem.edu}
\and
Yaw Adu-Gyamfi\\
Department of Civil Engineering\\
University of Missouri-Columbia\\
{\tt\small {adugyamfiy}@missouri.edu}
}
\maketitle

\begin{abstract}
Identifying unusual driving behaviors exhibited by drivers during driving is essential for understanding driver behavior and the underlying causes of crashes. Previous studies have primarily approached this problem as a classification task, assuming that naturalistic driving videos come discretized. However, both activity segmentation and classification are required for this task due to the continuous nature of naturalistic driving videos. The current study therefore departs from conventional approaches and introduces a novel methodological framework, \textbf{DeepSegmenter}, that simultaneously performs activity segmentation and classification in a single framework.  The proposed framework consists of four major modules namely Data Module, Activity Segmentation Module, Classification Module and Postprocessing Module. Our proposed method won \textbf{8th place} in the 2023 AI
City Challenge, Track 3, with an activity overlap score of \textbf{0.5426} on experimental validation data. The experimental results demonstrate the effectiveness, efficiency, and robustness of the proposed system. The code is available at \url{https://github.com/aboah1994/DeepSegment.git}.  

\end{abstract}

\section{Introduction}
\label{sec:intro}

Driving is a complex activity that necessitates a high level of concentration and coordination. Even with the best intentions, drivers may engage in behaviors that can result in accidents, such as distracted driving, aggressive driving, or driving while impaired. Our ability to identify and characterize these behaviors is essential for enhancing road safety and preventing traffic accidents. To do the aforementioned, a comprehensive dataset detailing different anomalous driving behaviors and differentiating them from safe driving actions is required.

\begin{figure}
    \centering
    \includegraphics[width=8cm]{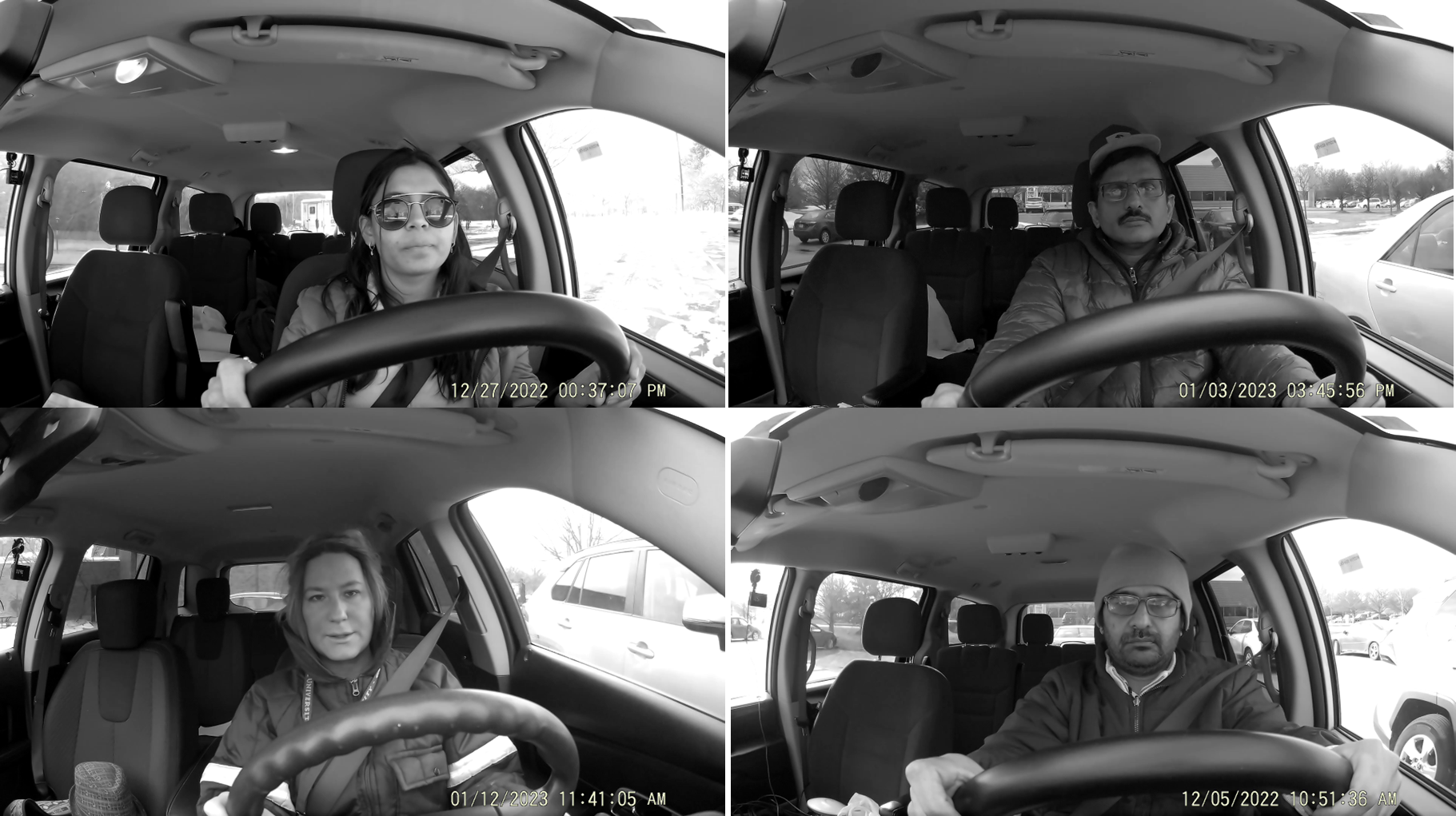}
    \caption{Cameras mounted on vehicle's dashboards to monitor drivers' behavior in a naturalistic environment.}
    \label{fig:driver}
\end{figure}
The naturalistic driving dataset is one such dataset that has been used extensively to study unsafe driving behaviors. As shown in Fig.~\ref{fig:driver}, naturalistic driving data refers to information collected from sensors or cameras mounted on vehicles as they are driven in real-world environments. This type of data contains a wealth of information about the driving task, including the actions of the driver and the context in which they are performed. This makes it possible to identify patterns and trends in naturalistic driving videos that may indicate the presence of anomalous or unsafe driving behaviors.

Driving activity recognition models usually capitalize on the richness of the visual feed provided by driver-facing camera data, to detect anomalous driving behaviors\cite{shaaban2021posture, braunagel2015driver, yang2021recognition,aboah2023driver,Aboah23AIC23}. The usual approach uses rule-based pose estimation and posture tracking, through feature detection and tracking algorithms\cite{shaaban2021posture, braunagel2015driver, martin2018body}. Other studies have focused on gaze mapping\cite{yang2019dual}, as well as the fusion of driver visual data with vehicle state characteristics\cite{yang2021identification}. This approach, whilst widely used and successful, only solves one part of the problem. Another limitation to this approach is the one-size-fits-all nature, which brings limited flexibility and adaptability to the defined rules governing the separation of normal and anomalous driving.

Given these limitations, advancements in anomalous driving behavior recognition and classification have tapped into the potential of more robust deep learning (DL) approaches\cite{yang2021identification, yang2021recognition}. Researchers favor deep learning approaches because they produce better predictions and outcomes than conventional rule based and machine learning algorithms. Deep learning models are mostly utilized to enable automatic feature extraction through the training of complicated features with little external assistance to provide meaningful representations of data through deep neural networks\cite{tanberk2020hybrid}.

Whilst DL approaches come with many advantages, the initial hurdle for getting them off the ground involves providing large amounts of pre-labeled data for their training\cite{gu2018ava, pham2022videobased,Naphade22AIC22}. Although this limitation is easily overcome with the availability of naturalistic driving data, the DL models are trained via sequence of event frames from video data. As such, these models expect pre-segmented video data clips in order to accurately classify the type of event taking place\cite{Chao_2018_CVPR}.

Video data streamed from cameras in a naturalistic driving environment are continuous and untrimmed in nature. In order to actively deploy activity recognition and classification models to such a situation, the modeling framework should have the ability to identify when anomalous driving activity is taking place whilst accurately classifying what kind of activity that is. This involves a summarized three step process of:
\begin{enumerate}
    \item Identifying and extracting visual features from video frames
    \item Defining the state of the driving behavior and observational period based on instances related to the extracted features
    \item Classifying the type of driving behavior observed in the localized time sequence where the action is observed
\end{enumerate}

As earlier highlighted, the more popular rule based models do a stellar job with steps 1 and 2, but are limited in their ability to accurately classify the type of observed action (step 3), especially when there are multiple classes of observed behaviors with little variations between groups of them\cite{yang2021recognition}. DL methods also have no trouble with steps 1 and 3 but usually require sequential or batch processing of the video frames\cite{Chao_2018_CVPR}. Whilst it’s possible to localize actions from temporal sequence by introducing techniques from rule-based algorithms, the research in this area is lacking.

The purpose of this study is to introduce a hybrid approach to continuous video activity recognition by capitalizing on the advantages of the two aforementioned approaches. The NVIDIA AI City 2023 challenge presented an opportunity to tackle this problem. In this study, we introduce DeepSegmenter, a temporal action localization algorithm which combines a DL feature detection algorithm and rule based feature tracking algorithm to first localize instances of anomalous driving behaviors (steps 1 and 2), before passing the clipped sequence of frames into a DL based activity classifier (step 3) to label the observed action as either normal driving or one of fifteen other anomalous driving behaviors.

DeepSegmenter performs incredibly well at localizing the instances of anomalous driving behaviors regardless of the length of the activity. The benefit of this approach is realized in its capacity to improve upon the current capabilities of Advanced Driver Assistance Systems (ADAS).

The rest of the paper is organized as follows. The second section discusses related research work on methods used in characterizing anomalous driving behaviour in  naturalistic driving videos. The third section presents our proposed approach, while the fourth section details the experiment procedure. We then present our experimental results, which demonstrate the effectiveness of our proposed method in segmenting and classifying anomaly detection. Lastly, we discuss the implications of our findings and propose future directions for research in this field.

\section{Related works}
\label{sec:lit}
Driver action recognition has been thoroughly explored in recent years, with research into it still ongoing. By creating models to recognize or forecast driving behaviors, and implementing remedial actions for anomalous and risky driving, it will be possible to improve the safety of vehicle driving and reduce the number of driver-caused road traffic accidents. Several research publications have employed various strategies to present solutions to this idea, with the most modern methodologies relying on supervised learning strategies.

\subsection{Traditional Methods}
Earlier research in human activity recognition relied on empirical rule inference and quantitative statistical analysis, such as Hidden Markov Model (HMM), Gaussian Mixture Model (GMM), Random Forest (RF), Support Vector Machine (SVM), Fuzzy Neural Network (FNN), and k-Nearest Neighbor (kNN) \cite{zhao2022review,aboah2020smartphone,aboah2021vision,boah2021mobile,aboah2021comparative}. Using a random forest classification algorithm, Ahnagari et al. \cite{ahangari2021enhancing} identified six prevalent distracted driving behaviors with a 0.765 accuracy rate. Yao et al. \cite{yao2018classification} also developed a random forest model to identify distracted driving behavior with 0.9 accuracy. Whilst these accuracies suggest incredible strides, they are limited to binary outcomes of anomalous and normal driving, and fail in multi-classification tasks of identifying the types of anomalous driving activity observed. Traditional methods have provided us with a reasonable degree of accuracy, but their drawbacks, such as dependence on specialist experience on artificial extraction of characteristics, and inability to consider driving time sequence and correlation, can lead to driving behavior identification errors \cite{zhao2022review,aboah2021identifying}. More recent advancements in the area of driver action recognition have looked into bridging this limitation by utilizing more advanced deep learning based models.

\subsection{Deep learning Methods}
The deep spatiotemporal features of driving behavior data can be automatically extracted using deep learning techniques like Convolutional Neural Networks (CNN), Recurrent Neural Networks (RNN), as well as Transformers. These techniques also incorporate feature extraction and recognition prediction into a model for end-to-end learning with high recognition accuracy. Current temporal action detection models can be classified into either single-stage detectors or two-stage detectors.

\vspace{0.1in}
\noindent\textbf{Single-Stage Detector}. Single-stage detectors predict action proposals directly from video features, without the need for intermediate processes like region proposals or sliding windows \cite{tang2019afo, liu2022end, rahman2020single, cheng2021multi}. Anchor Free single-stage detector employed by Tang et al. \cite{tang2019afo} predicts the action boundaries and scores directly from the video features using an anchor free regression head and a focal loss function. Decouple-SSAD method was proposed by Rahman and Laganiere \cite{rahman2020single}, this method decouples the localization and classification tasks by using parallel branches and a soft-NMS module to refine the proposals.  Chen et al. \cite{chen2021timeception} offers a single-stage approach that makes use of Timeception (temporal CNN), which collects multi-scale temporal information and produces action proposals and classification. Yan et al. \cite{yan2016driving} pre-trained a CNN model using unsupervised feature learning via sparse filtering, followed by fine-tuning with classification. Their system monitors the position of the
driver’s hands and uses extracted information to predict whether the posture is safe or unsafe. Baheti et al. \cite{baheti2020towards} also used a MobileVGG network  to detect and classify driver distraction. With the rise of the Transformer's attention technique that has achieved state-of-the-art results, many recent research solutions have implemented transformers in temporal action detection \cite{liu2022end,li2022mv,zhang2022actionformer}.

\vspace{0.1in}
\noindent\textbf{Two-Stage Detector}. A two-stage detector has two stages: one for extracting frames and one for classifying proposals/redefining temporal boundaries. Inspired by Faster-RCNN, Chao et al. \cite{chao2018rethinking}, as well as Gao et al. \cite{gao2017turn}, used a multi-scale architecture to improve receptive field alignment and exploits the temporal context of actions for proposal generation and action classification. ActionFormer, initially developed by Zhang et al.\cite{zhang2022actionformer}, was employed by Nguyen \cite{nguyen2022learning} to predict the temporal location of an event and do classification simultaneously, a second-stage classifier is then employed to improve prediction precision. Stragazer is also another efficient multi-scale vision transformer that learns hierarchical robust representations and then uses a sliding window for temporal localization\cite{liang2022stargazer}. Ding et al.\cite{ding2022coarse} also proposed a Coarse-to-Fine Boundary Localization Method in which the features of the video are extracted first, and then a sliding window is used to generate coarse boundaries. Following that, the boundaries are refined to obtain the fine boundaries. 

\section{Approach}
\label{sec:meth}
\begin{figure*}[t!]
    \centering
    \includegraphics[width=17cm]{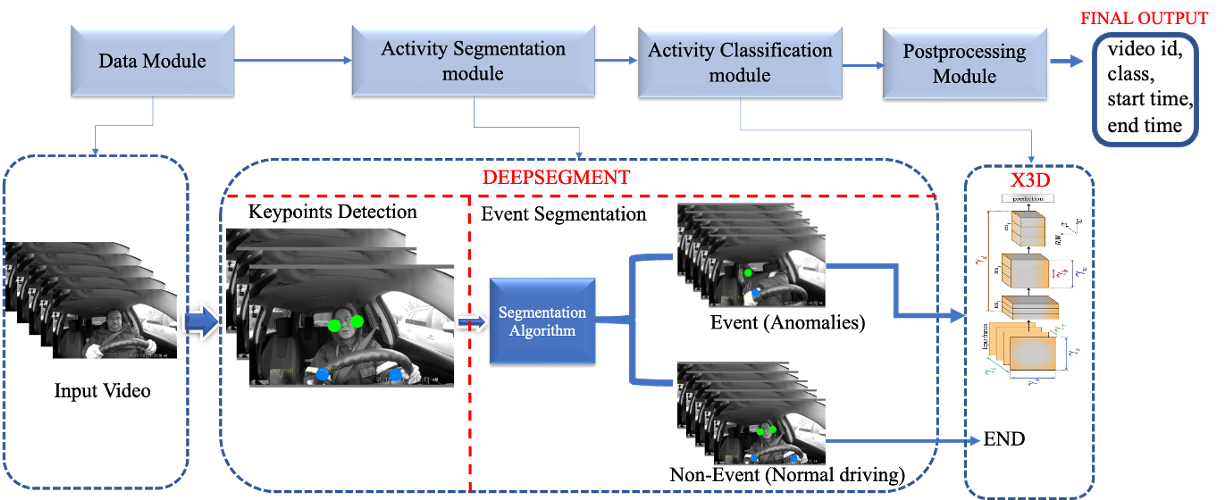}
    \caption{Overall Structure of DeepSegmenter.}
    \label{fig:pred_img}
\end{figure*}

The overall structure of our DeepSegmenter system is illustrated in Fig~\ref{fig:pred_img}. Our design is simple yet effective. The proposed system consists of four primary modules: Data Module, Activity Segmentation Module, Classification Module and Postprocessing Module.

\subsection{Data Module}
The data module is tasked with organizing and preprocessing the naturalistic driving videos. To accomplish this, the data module executes a series of preprocessing operations, including resolution normalization and pixel-value normalization. Resolution normalization entails resizing the video to a standard resolution, which ensures that the video can be analyzed uniformly across all samples as shown in Equation~\ref{eq:res}. Pixel-value normalization involves rescaling the pixel values to a standard range of 0 and 1, which eliminates differences in brightness and contrast between videos (Equation~\ref{eq:res1}). By performing these preprocessing steps, the data module ensures that the collected data is in a format that subsequent modules can easily process and analyze.

\begin{equation}
\label{eq:res}
X_{resized} = \text{resize}(X_{original}, (W, H))
\end{equation}

where $X_{original}$ is the original video, $X_{resized}$ is the resized video, and $\text{resize}()$ is a function that resizes the video to the specified dimensions.

\begin{equation}
\label{eq:res1}
X_{normalized} = \frac{X_{resized} - X_{min}}{X_{max} - X_{min}}
\end{equation}

where $X_{resized}$ is the resized video, $X_{min}$ is the minimum pixel value in $X_{resized}$, $X_{max}$ is the maximum pixel value in $X_{resized}$, and $X_{normalized}$ is the normalized video.

\subsection{Activity Segmentation Module}
The activity segmentation module breaks down the continuous stream of video data into discrete segments that can be analyzed and later classified. This module consists of two submodules:

\vspace{0.1in}
\noindent\textbf{Keypoint Detection}. The keypoint detection submodule identifies key points in each frame of the video, such as the face and hands of the driver, which are required by the activity segmentation step. This submodule employs a pretrained yolov7\cite{wang2022yolov7} keypoint detections model to detect and track the movement of these keypoints across multiple frames. 

\vspace{0.1in}
\noindent\textbf{Activity Segmentation}. The activity segmentation submodule uses detected key points to identify and classify driver activities as either event (anomaly) or non-event (normal driving). This submodule utilizes heuristic-based algorithm for it categorization as illustrated in Algorithm~\ref{alg:activitysegment}. Anomalies are triggered by either hand or head movements of the driver. In the case of a head anomaly, the submodule detects when the angle between the eyes and the nose surpasses a predefined threshold, if so, it classifies the frame as an anomaly, otherwise, it is considered as normal driving. Similarly, for a hand anomaly, when the angle between the hand exceeds a predefined threshold, an anomaly is triggered, else it is classified as normal driving.


\begin{algorithm}
\caption{Activity Segmentation}
\label{alg:activitysegment}

\textbf{Require:} $\theta_h, \theta_{hand}$ \textit{(threshold angles for head and hand movements)}

\textbf{Ensure:} \textit{classification result} $c$ \textit{for each frame}

\begin{enumerate}
\item \textbf{Function} \textsc{classifyActivity}(\textit{frame}):
    \begin{enumerate}
    \item \textit{keyPoints} $\gets$ \textsc{extractKeyPoints}(frame)
    \item \textit{headAngle} $\gets$ \textsc{calculateHeadAngle}(keyPoints)
    \item \textit{handAngle} $\gets$ \textsc{calculateHandAngle}(keyPoints)
    \item \textbf{if} headAngle $>$ $\theta_h$ \textbf{then}
    \item \hspace{0.5cm} \textbf{return} \textit{Anomaly}
    \item \textbf{else if} handAngle $>$ $\theta_{hand}$ \textbf{then}
    \item \hspace{0.5cm} \textbf{return} \textit{Anomaly}
    \item \textbf{else}
    \item \hspace{0.5cm} \textbf{return} \textit{Normal Driving}
    \end{enumerate}

\item \textbf{Function} \textsc{Main}():
    \begin{enumerate}
    \item \textbf{for each} frame $f$ \textbf{do}
    \item \hspace{0.5cm} \textit{classificationResult} $\gets$ \textsc{classifyActivity}($f$)
    \item \hspace{0.5cm} \textsc{storeResult}(\textit{classificationResult})
    \end{enumerate}
\end{enumerate}
\end{algorithm}

\subsection{Activity Classification Module}
The activity classification module is in charge of classifying the segmented event into 1 of 15 different categories of driving anomalies as shown in Table~\ref{tab:my-table1}. This module uses a 3D CNN architecture known as X3D\cite{feichtenhofer2020x3d}, developed for video analysis. 

\begin{figure}[H]
    \centering
    \includegraphics[width=8cm]{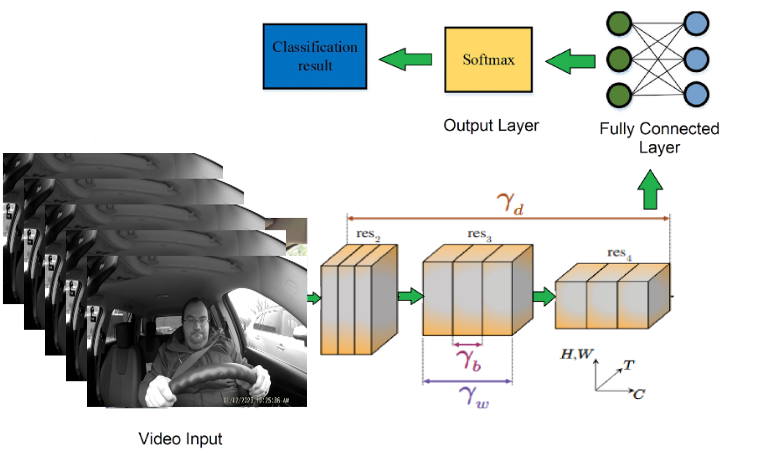}
    \caption{X3D model architecture.}
    \label{fig:3xd}
\end{figure}

\vspace{0.1in}
\noindent\textbf{X3D Architecture}. The X3D structure is designed to efficiently process video data using a combination of 2D and 3D convolutions. The central idea behind X3D is to expand the network in both spatial and temporal dimensions to allow for greater expressiveness while preserving efficiency. The three components of the X3D network are the entry flow, the middle flow, and the exit flow. The entry flow initially processes incoming video frames using a series of 2D convolutional layers, followed by a 3D convolutional layer containing temporal information. The middle flow adds additional 3D convolutional layers to the network. The exit flow then reduces the output's spatial dimensions using a combination of 2D and 3D convolutions making it easier for classification by fully connected layer as demonstrated in Fig~\ref{fig:3xd}.

X3D networks enlarge a 2D network along multiple axes, such as duration, frame rate, spatial resolution, width, bottleneck width, and depth. The application of channel-wise spatio-temporal convolutions that enable the efficient processing of video data is one of the most significant developments of X3D. These convolutions make use of shared weights across neighboring channels in a feature map, thereby reducing the number of required parameters and increasing efficiency. The "factorized" design of X3D, in which 3D convolutions are broken down into a series of 2D convolutions, reduces the computational cost of 3D convolutions while preserving their ability to capture temporal information.

\subsection{Postprocessing Module}
The postprocessing module removes false positive detections, which consist primarily of events extracted for less than one second. It employs a rule-based algorithm that estimates the duration of classified events and either ignores or adds to the final submission if the duration exceeds 1 second. Finally, this module prepares the final results in the format required by the evaluation system.





\section{Experiment}
In this study, we assess the effectiveness of the DeepSegmenter system on the AI City Challenge dataset for Naturalistic Driver Action Recognition. Our findings reveal that our model performs well compared to other systems when tackling this difficult task.

\vspace{0.1in}
\noindent\textbf{Data}. The data set consists of 210 video clips that amount to approximately 34 hours of footage, which were taken from 35 different drivers\cite{Naphade23AIC23,Rahman22SynDD2}. Each driver performed 16 different tasks, including actions like talking on the phone, eating, and reaching back, once and in a random order. Each vehicle was equipped with three cameras that recorded from different angles in synchronization. To collect the data, each driver completed the tasks twice: once without any appearance block and once while wearing an appearance block such as sunglasses or a hat. This resulted in a total of 6 videos per each driver, with 3 videos recorded without an appearance block and 3 videos recorded with an appearance block. The summary of the activities performed by the drivers are summarized in Table~\ref{tab:my-table1}.

\begin{table}[h]
\caption{Driver Activity Summary}
\label{tab:my-table1}
\begin{tabular}{@{}ll@{}}
\toprule
Activity Class type & Activity Label                 \\ \midrule
1                   & Drinking                       \\
2                   & Phone Call(right)              \\
3                   & Phone Call(left)               \\
4                   & Eating                         \\
5                   & Text (Right)                   \\
6                   & Text (Left)                    \\
7                   & Reaching behind                \\
8                   & Adjust control panel           \\
9                   & Pick up from floor (Driver)    \\
10                  & Pick up from floor (Passenger) \\
11                  & Talk to passenger at the right \\
12                  & Talk to passenger at backseat  \\
13                  & yawning                        \\
14                  & Hand on head                   \\
15                  & Singing or dancing with music  \\ \bottomrule
\end{tabular}
\end{table}

\vspace{0.1in}
\noindent\textbf{Training}. The experiments were carried out on an NVIDIA GTX 1080ti GPU graphics card. The activity classification model was built on the PyTorch Lightning framework. The dataset was partitioned into ratios of 0.7:0.3, corresponding to the training and validation datasets, respectively. We used the Adam optimizer with a starting learning rate of 0.001 and a weight decay of 0.001. We used the CosineAnnealingLR scheduler to adjust the learning rate during training. The model was trained for 300 epochs with a batch size of 8.

\vspace{0.1in}
\noindent\textbf{Evaluation metrics}. The average activity overlap score was used as the evaluation metric in this study as shown in Equation~\ref{eq:1}. This score is determined by comparing the predicted activity with the ground-truth activity based on their overlap. The closest match will be considered as the predicted activity with the highest overlap score \textit{(os)}, provided that it belongs to the same class as the ground-truth activity. However, this match will only be considered if the predicted activity's start time \textit{(ps)} and end time \textit{(pe)} are within a range of 10 seconds before or after the ground-truth activity's start time \textit{(gs)} and end time \textit{(ge)}, respectively. The overlap score is computed by finding the ratio between the time intersection and time union of the two activities.

\begin{equation}
\label{eq:1}
o s(p, g)=\frac{\max (\min (g e, p e)-\max (g s, p s), 0)}{\max (g e, p e)-\min (g s, p s)}
\end{equation}



\section{Results and Discussion}
\label{sec:res}
The 2023 NVIDIA AI City Challenge Track 3 test videos consists of 30 untrimmed videos from 5 different drivers at 3 different camera positions. A submission to the competition is a text file that follows the format: Video ID, Activity ID, Start time, and End time. The Video ID is a numeric identifier for the video, starting with 1 and indicating its position in the alphabetically ordered list of all Track 3, Test Set videos.  The Activity ID is a numeric identifier beginning with 1 for the classified class of the anomaly.The Start and End times are integer values representing the beginning and end of the anomalous activity, respectively.

On the experimental test dataset, our proposed methodology achieved an overall overlap score of \textbf{0.5426}, ranking \textbf{8th} on the public leader board, as shown in Table~\ref{tab:my-table2}.

\begin{table}[h]
\centering
\caption{Top 10 Leader Board Ranking}
\label{tab:my-table2}
\begin{tabular}{@{}llll@{}}
\toprule
\textbf{Rank}   & Team ID  & Team Name & Score    \\ \midrule
1 & 209 & Meituan\-IoTCV & 0.7416          \\
2 & 60 & JNU\_boat & 0.7041          \\
3 & 49 & ctc\-AI & 0.6723          \\
4 & 118 & RW & 0.6245   \\
5 & 8 & Purdue Digital Twin & 0.5921         \\
6 & 48 & BUPT\-MCPRL & 0.5907         \\
7 & 83 & DiveDeeper & 0.5881         \\
\textcolor{red}{8} & \textcolor{red}{217} & \textcolor{red}{INTELLI\_LAB (Ours)} & \textcolor{red}{0.5426}          \\
9 & 152 & AI\_LAB & 0.5424          \\
10 & 11 & AIMIZ & 0.5409          \\

 \bottomrule
\end{tabular}%
\end{table}

\section{Conclusion}
\label{sec:con}
This study presents a solution for Track 3 of the 2023 AI City Challenge that focuses on performing both activity segmentation and classification in a single framework called DeepSegmenter. The proposed framework is composed of four modules: Data Module, Activity Segmentation Module, Classification Module, and Postprocessing Module. According to the experimental results on the test dataset, the proposed framework ranks 8th in the challenge with an overlap score of 0.5426. On this challenge, we demonstrated the effectiveness of our proposed framework.


{\small
\bibliographystyle{unsrt}
\bibliography{egbib}

\begin{thebibliography}{10}

\bibitem{shaaban2021posture}
Khaled Shaaban, Osama Halabi, and Eman Almughani.
\newblock A posture recognition system to track drivers’ activities while
  driving.
\newblock {\em Journal of Traffic and Transportation Management}, 3(2):25--31,
  2021.

\bibitem{braunagel2015driver}
Christian Braunagel, Enkelejda Kasneci, Wolfgang Stolzmann, and Wolfgang
  Rosenstiel.
\newblock Driver-activity recognition in the context of conditionally
  autonomous driving.
\newblock In {\em 2015 IEEE 18th International Conference on Intelligent
  Transportation Systems}, pages 1652--1657. IEEE, 2015.

\bibitem{yang2021recognition}
Lichao Yang, Kuo Dong, Yan Ding, James Brighton, Zhenfei Zhan, and Yifan Zhao.
\newblock Recognition of visual-related non-driving activities using a
  dual-camera monitoring system.
\newblock {\em Pattern Recognition}, 116:107955, 2021.

\bibitem{aboah2023driver}
Armstrong Aboah, Yaw Adu-Gyamfi, Senem~Velipasalar Gursoy, Jennifer Merickel,
  Matt Rizzo, and Anuj Sharma.
\newblock Driver maneuver detection and analysis using time series segmentation
  and classification.
\newblock {\em Journal of Transportation Engineering, Part A: Systems},
  149(3):04022157, 2023.

\bibitem{Aboah23AIC23}
Armstrong Aboah, Bin Wang, Bagci Ulas, and Yaw Adu-Gyamfi.
\newblock Real-time multi-class helmet violation detection using few-shot data
  sampling technique and yolov8.
\newblock In {\em The IEEE Conference on Computer Vision and Pattern
  Recognition (CVPR) Workshops}, June 2023.

\bibitem{martin2018body}
Manuel Martin, Johannes Popp, Mathias Anneken, Michael Voit, and Rainer
  Stiefelhagen.
\newblock Body pose and context information for driver secondary task
  detection.
\newblock In {\em 2018 IEEE Intelligent Vehicles Symposium (IV)}, pages
  2015--2021. IEEE, 2018.

\bibitem{yang2019dual}
Lichao Yang, Kuo Dong, Arkadiusz~Jan Dmitruk, James Brighton, and Yifan Zhao.
\newblock A dual-cameras-based driver gaze mapping system with an application
  on non-driving activities monitoring.
\newblock {\em IEEE Transactions on Intelligent Transportation Systems},
  21(10):4318--4327, 2019.

\bibitem{yang2021identification}
Lichao Yang, Mahdi Babayi~Semiromi, Yang Xing, Chen Lv, James Brighton, and
  Yifan Zhao.
\newblock The identification of non-driving activities with associated
  implication on the take-over process.
\newblock {\em Sensors}, 22(1):42, 2021.

\bibitem{tanberk2020hybrid}
Senem Tanberk, Zeynep~Hilal Kilimci, Dilek~Bilgin T{\"u}kel, Mitat Uysal, and
  Selim Akyoku{\c{s}}.
\newblock A hybrid deep model using deep learning and dense optical flow
  approaches for human activity recognition.
\newblock {\em IEEE Access}, 8:19799--19809, 2020.

\bibitem{gu2018ava}
Chunhui Gu, Chen Sun, David~A. Ross, Carl Vondrick, Caroline Pantofaru, Yeqing
  Li, Sudheendra Vijayanarasimhan, George Toderici, Susanna Ricco, Rahul
  Sukthankar, Cordelia Schmid, and Jitendra Malik.
\newblock Ava: A video dataset of spatio-temporally localized atomic visual
  actions, 2018.

\bibitem{pham2022videobased}
Hieu~H. Pham, Louahdi Khoudour, Alain Crouzil, Pablo Zegers, and Sergio~A.
  Velastin.
\newblock Video-based human action recognition using deep learning: A review,
  2022.

\bibitem{Naphade22AIC22}
M.~Naphade, S.~Wang, D.~C. Anastasiu, Z.~Tang, M.~Chang, Y.~Yao, L.~Zheng,
  M.~Shaiqur Rahman, A.~Venkatachalapathy, A.~Sharma, Q.~Feng, V.~Ablavsky,
  S.~Sclaroff, P.~Chakraborty, A.~Li, S.~Li, and R.~Chellappa.
\newblock The 6th ai city challenge.
\newblock In {\em 2022 IEEE/CVF Conference on Computer Vision and Pattern
  Recognition Workshops (CVPRW)}, pages 3346--3355. IEEE Computer Society, June
  2022.

\bibitem{Chao_2018_CVPR}
Yu-Wei Chao, Sudheendra Vijayanarasimhan, Bryan Seybold, David~A. Ross, Jia
  Deng, and Rahul Sukthankar.
\newblock Rethinking the faster r-cnn architecture for temporal action
  localization.
\newblock In {\em Proceedings of the IEEE Conference on Computer Vision and
  Pattern Recognition (CVPR)}, June 2018.

\bibitem{zhao2022review}
Dengfeng Zhao, Yudong Zhong, Zhijun Fu, Junjian Hou, Mingyuan Zhao, et~al.
\newblock A review for the driving behavior recognition methods based on
  vehicle multisensor information.
\newblock {\em Journal of Advanced Transportation}, 2022, 2022.

\bibitem{aboah2020smartphone}
Armstrong Aboah and Yaw Adu-Gyamfi.
\newblock Smartphone-based pavement roughness estimation using deep learning
  with entity embedding.
\newblock {\em Advances in Data Science and Adaptive Analysis},
  12(03n04):2050007, 2020.

\bibitem{aboah2021vision}
Armstrong Aboah.
\newblock A vision-based system for traffic anomaly detection using deep
  learning and decision trees.
\newblock In {\em Proceedings of the IEEE/CVF Conference on Computer Vision and
  Pattern Recognition}, pages 4207--4212, 2021.

\bibitem{boah2021mobile}
Armstrong Aboah, Michael Boeding, and Yaw Adu-Gyamfi.
\newblock Mobile sensing for multipurpose applications in transportation.
\newblock {\em arXiv preprint arXiv:2106.10733}, 2021.

\bibitem{aboah2021comparative}
Armstrong Aboah and Elizabeth Arthur.
\newblock Comparative analysis of machine learning models for predicting travel
  time.
\newblock {\em arXiv preprint arXiv:2111.08226}, 2021.

\bibitem{ahangari2021enhancing}
Samira Ahangari, Mansoureh Jeihani, Anam Ardeshiri, Md~Mahmudur Rahman, and
  Abdollah Dehzangi.
\newblock Enhancing the performance of a model to predict driving distraction
  with the random forest classifier.
\newblock {\em Transportation research record}, 2675(11):612--622, 2021.

\bibitem{yao2018classification}
Ying Yao, Xiaohua Zhao, Hongji Du, Yunlong Zhang, and Jian Rong.
\newblock Classification of distracted driving based on visual features and
  behavior data using a random forest method.
\newblock {\em Transportation research record}, 2672(45):210--221, 2018.

\bibitem{aboah2021identifying}
Armstrong Aboah, Lydia Johnson, and Setul Shah.
\newblock Identifying the factors that influence urban public transit demand.
\newblock {\em arXiv preprint arXiv:2111.09126}, 2021.

\bibitem{tang2019afo}
Yiping Tang, Chuang Niu, Minghao Dong, Shenghan Ren, and Jimin Liang.
\newblock Afo-tad: Anchor-free one-stage detector for temporal action
  detection.
\newblock {\em arXiv preprint arXiv:1910.08250}, 2019.

\bibitem{liu2022end}
Xiaolong Liu, Qimeng Wang, Yao Hu, Xu~Tang, Shiwei Zhang, Song Bai, and Xiang
  Bai.
\newblock End-to-end temporal action detection with transformer.
\newblock {\em IEEE Transactions on Image Processing}, 31:5427--5441, 2022.

\bibitem{rahman2020single}
Md~Atiqur Rahman and Robert Lagani{\`e}re.
\newblock Single-stage end-to-end temporal activity detection in untrimmed
  videos.
\newblock In {\em 2020 17th Conference on Computer and Robot Vision (CRV)},
  pages 206--213. IEEE, 2020.

\bibitem{cheng2021multi}
Rao Cheng, Xiaowei He, Zhonglong Zheng, and Zhentao Wang.
\newblock Multi-scale safety helmet detection based on sas-yolov3-tiny.
\newblock {\em Applied Sciences}, 11(8):3652, 2021.

\bibitem{chen2021timeception}
Xiaoqiu Chen, Miao Ma, Zhuoyu Tian, and Jie Ren.
\newblock Timeception single shot action detector: A single-stage method for
  temporal action detection.
\newblock In {\em Image and Graphics: 11th International Conference, ICIG 2021,
  Haikou, China, August 6--8, 2021, Proceedings, Part I 11}, pages 340--354.
  Springer, 2021.

\bibitem{yan2016driving}
Chao Yan, Frans Coenen, and Bailing Zhang.
\newblock Driving posture recognition by convolutional neural networks.
\newblock {\em IET Computer Vision}, 10(2):103--114, 2016.

\bibitem{baheti2020towards}
Bhakti Baheti, Sanjay Talbar, and Suhas Gajre.
\newblock Towards computationally efficient and realtime distracted driver
  detection with mobilevgg network.
\newblock {\em IEEE Transactions on Intelligent Vehicles}, 5(4):565--574, 2020.

\bibitem{li2022mv}
Wei Li, Shimin Chen, Jianyang Gu, Ning Wang, Chen Chen, and Yandong Guo.
\newblock Mv-tal: Mulit-view temporal action localization in naturalistic
  driving.
\newblock In {\em Proceedings of the IEEE/CVF Conference on Computer Vision and
  Pattern Recognition}, pages 3242--3248, 2022.

\bibitem{zhang2022actionformer}
Chen-Lin Zhang, Jianxin Wu, and Yin Li.
\newblock Actionformer: Localizing moments of actions with transformers.
\newblock In {\em Computer Vision--ECCV 2022: 17th European Conference, Tel
  Aviv, Israel, October 23--27, 2022, Proceedings, Part IV}, pages 492--510.
  Springer, 2022.

\bibitem{chao2018rethinking}
Yu-Wei Chao, Sudheendra Vijayanarasimhan, Bryan Seybold, David~A Ross, Jia
  Deng, and Rahul Sukthankar.
\newblock Rethinking the faster r-cnn architecture for temporal action
  localization.
\newblock In {\em Proceedings of the IEEE conference on computer vision and
  pattern recognition}, pages 1130--1139, 2018.

\bibitem{gao2017turn}
Jiyang Gao, Zhenheng Yang, Kan Chen, Chen Sun, and Ram Nevatia.
\newblock Turn tap: Temporal unit regression network for temporal action
  proposals.
\newblock In {\em Proceedings of the IEEE international conference on computer
  vision}, pages 3628--3636, 2017.

\bibitem{nguyen2022learning}
Chuong Nguyen, Ngoc Nguyen, Su~Huynh, Vinh Nguyen, and Son Nguyen.
\newblock Learning generalized feature for temporal action detection:
  Application for natural driving action recognition challenge.
\newblock In {\em Proceedings of the IEEE/CVF Conference on Computer Vision and
  Pattern Recognition}, pages 3249--3256, 2022.

\bibitem{liang2022stargazer}
Junwei Liang, He~Zhu, Enwei Zhang, and Jun Zhang.
\newblock Stargazer: A transformer-based driver action detection system for
  intelligent transportation.
\newblock In {\em Proceedings of the IEEE/CVF Conference on Computer Vision and
  Pattern Recognition}, pages 3160--3167, 2022.

\bibitem{ding2022coarse}
Guanchen Ding, Wenwei Han, Chenglong Wang, Mingpeng Cui, Lin Zhou, Dianbo Pan,
  Jiayi Wang, Junxi Zhang, and Zhenzhong Chen.
\newblock A coarse-to-fine boundary localization method for naturalistic
  driving action recognition.
\newblock In {\em Proceedings of the IEEE/CVF Conference on Computer Vision and
  Pattern Recognition}, pages 3234--3241, 2022.

\bibitem{wang2022yolov7}
Chien-Yao Wang, Alexey Bochkovskiy, and Hong-Yuan~Mark Liao.
\newblock Yolov7: Trainable bag-of-freebies sets new state-of-the-art for
  real-time object detectors.
\newblock {\em arXiv preprint arXiv:2207.02696}, 2022.

\bibitem{feichtenhofer2020x3d}
Christoph Feichtenhofer.
\newblock X3d: Expanding architectures for efficient video recognition.
\newblock In {\em Proceedings of the IEEE/CVF conference on computer vision and
  pattern recognition}, pages 203--213, 2020.

\bibitem{Naphade23AIC23}
Milind Naphade, Shuo Wang, David~C. Anastasiu, Zheng Tang, Ming-Ching Chang,
  Yue Yao, Liang Zheng, Mohammed~Shaiqur Rahman, Meenakshi~S. Arya, Anuj
  Sharma, Qi~Feng, Vitaly Ablavsky, Stan Sclaroff, Pranamesh Chakraborty,
  Sanjita Prajapati, Alice Li, Shangru Li, Krishna Kunadharaju, Shenxin Jiang,
  and Rama Chellappa.
\newblock The 7th {AI City Challenge}.
\newblock In {\em The IEEE Conference on Computer Vision and Pattern
  Recognition (CVPR) Workshops}, June 2023.

\bibitem{Rahman22SynDD2}
Mohammed~Shaiqur Rahman, Jiyang Wang, Senem~Velipasalar Gursoy, David
  Anastasiu, Shuo Wang, and Anuj Sharma.
\newblock {S}ynthetic {D}istracted {D}riving ({SynDD2}) dataset for analyzing
  distracted behaviors and various gaze zones of a driver, 2022.
\newblock arXiv:2204.08096.

\end{thebibliography}
}

\end{document}